\documentclass[letter, 10pt]{IEEEtran}
\usepackage{graphicx}
\usepackage{multicol}
\usepackage{subfig}
\usepackage{amsmath,amssymb}
\usepackage{multirow}

\usepackage[font=small]{caption}
\usepackage{amsmath}
\usepackage{amssymb}
\usepackage{float}

\usepackage{lettrine}
\usepackage{algorithm}
\usepackage{algpseudocode}
\usepackage{cite}
\usepackage{filecontents}
\usepackage{lipsum}
\usepackage{color}
\usepackage{esdiff}
\usepackage{epstopdf}
\usepackage[utf8]{inputenc}
\usepackage[english]{babel}
\usepackage{amsthm}
\usepackage{ulem}

\begin{document}

\title{
\huge {
MRRT: Multiple Rapidly-Exploring Random Trees for Fast Online Replanning in Dynamic Environments
}\\
}

\author{ \begin{tabular}{cccccccccc}
{Zongyuan Shen$^\dag$$^\star$} & {James P. Wilson$^\dag$} & {Ryan Harvey$^\dag$} & {Shalabh Gupta$^\dag$}
\end{tabular}\vspace{12pt}
\thanks {$^\dag$Dept. of Electrical and Computer Engineering, University of Connecticut, Storrs, CT 06269, USA.}
\thanks {$^\star$Corresponding author (email: zongyuan.shen@uconn.edu)}
\begin{center}
{\large \textbf{Abstract}} \vspace{-24pt}
\end{center}
}

\maketitle

\IEEEpeerreviewmaketitle

\thispagestyle{empty}

Motion planning is a fundamental task of determining a collision-free trajectory that drives the robot to a target state in the  configuration space, minimizing a user-defined cost function such as trajectory length\cite{song2019}. It has a wide range of applications for autonomous underwater vehicles (AUVs), such as seabed mapping\cite{shen2020ct}, structural inspection\cite{englot2013}, oil spill cleaning\cite{song2013}, mine hunting\cite{MGR11}, and other underwater tasks\cite{song2018}. In order to ensure that the AUV successfully completes its mission, the path planner must be adaptive to previously-unknown obstacles in the environment as well as moving obstacles (e.g., other marine vessels) \cite{Singh2018replan}. As such, it is critical to ensure rapid replanning of the AUV paths online to facilitate safe operation as new information becomes available.

A variety of methods have been developed to solve the motion planning problem; a review is presented in \cite{ZENG2015_AUVPathPlanningSurvey}. In general, motion planning methods can be broadly classified into two main categories: grid-based and sample-based. The grid-based methods, such as A$^\star$\cite{hart1968formal}, discretize the configuration space and search for the optimal solution; however, these approaches suffer from the curse of dimensionality and their accuracy depends on the grid resolution.  
On the other hand, the sampling-based methods generate samples randomly in the configuration space, construct a graph structure to capture connectivity between different configurations, and search a solution on it \cite{Elbanhawi2014_samplereview}. Therefore, sample-based motion planning approaches are becoming increasingly popular since they can find feasible solutions quickly in high-dimensional spaces while asymptotically approaching the optimal one. Specifically, they are very useful for online planning in dynamic environments. Online methods are characterized as active or reactive. Active strategies predict the future trajectory of a moving obstacle for a fixed time duration and generate a collision-free trajectory for the robot; however, the performance can degrade if the predicted information is incomplete or incorrect. In contrast, the reactive strategies plan the path based only on the obstacle information at the current time, and replan the path whenever the obstacle information changes.

Recently, several variants of the RRT/RRT$^\star$ have been presented for online motion planning in dynamic environments. Otte et al. \cite{otte2016rrtx} proposed RRT$^\text{X}$ algorithm which utilizes a rapidly-exploring random graph (RRG)\cite{karaman2011sampling} to explore the search area. This method maintains a well-defined and connected structure over the entire explored region; however, re-optimizing the connections in the entire graph when the environment changes makes it inefficient in highly dynamic settings. Chen et al.\cite{chen2019horizon} presented Horizon-based Lazy RRT$^\star$ (HL-RRT$^\star$) algorithm in the context of RRT$^\star$ algorithm\cite{karaman2011sampling}. In this method, invalid parts of the search tree due to moving obstacles are pruned, and new samples are drawn to find a new path using a trained Gaussian mixture model. While this can find a new solution quickly, the machine learning based sampler cannot guarantee the performance. Furthermore, this method removes the invalid parts which are disconnected from the main tree to maintain a single tree structure, resulting in repeated exploration of the same area.  Finally, Yuan et al.\cite{yuan2020efficient} proposed Efficient Bias-goal Factor RRT (EBG-RRT), where invalid parts of the tree are pruned if it does not contain the current position or destination. Then, leaf nodes are identified in the main tree that are near the destination tree. Finally, samples are drawn using heuristics or a uniform distribution to attempt to reconnect the trees. This method is easy to implement; however, it prunes the tree portions which do not contain the current node or goal, resulting in repeated exploration of the same area.

\begin{figure*}[t]
    \centering
    \subfloat[Obstacle information is updated; infeasible edges are pruned from the tree.]{
        \includegraphics[width=0.44\textwidth]{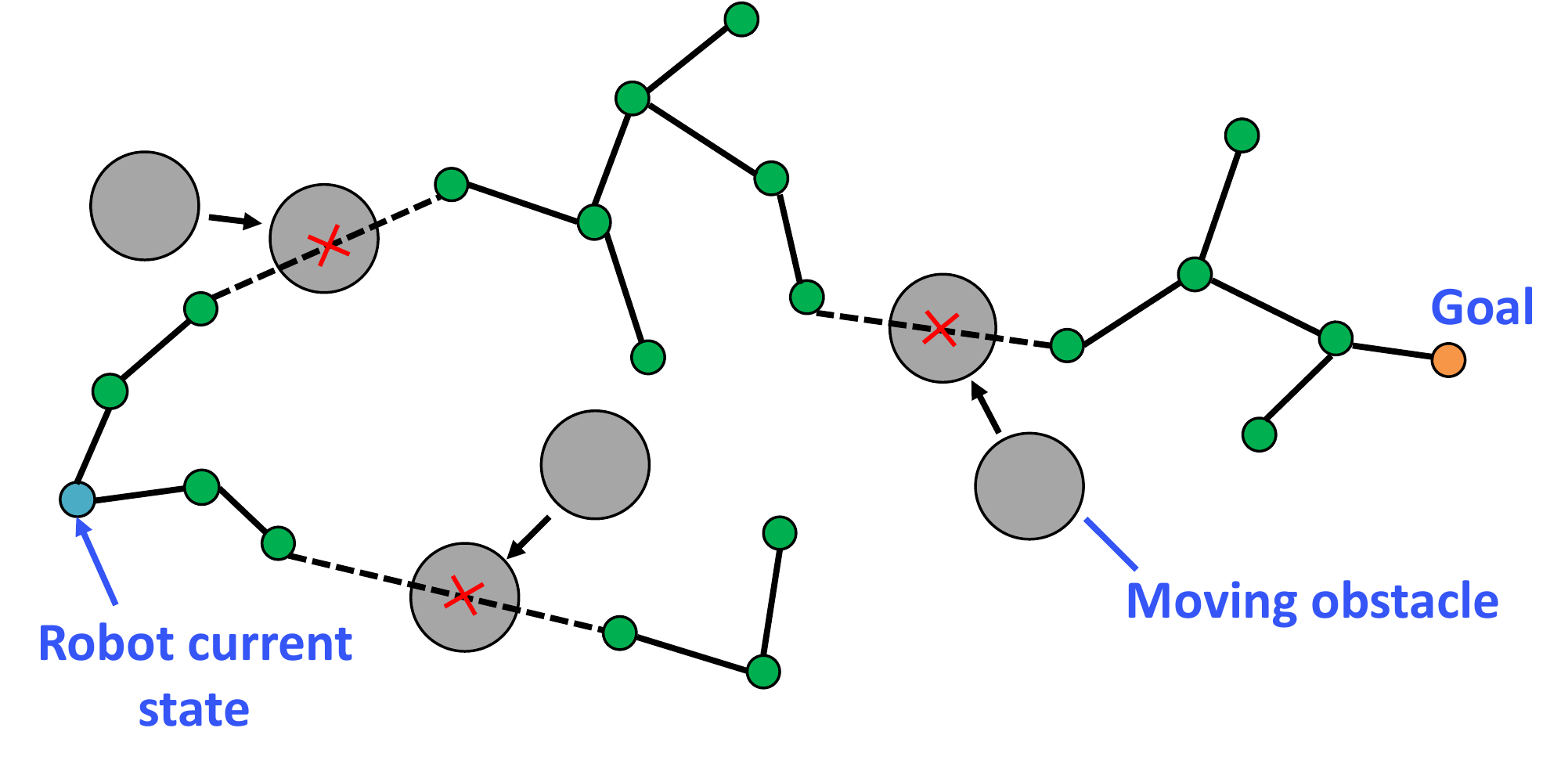}\label{fig:part1}}\quad
         \centering
    \subfloat[By removing infeasible edges multiple disjoint trees are created.]{
         \includegraphics[width=0.44\textwidth]{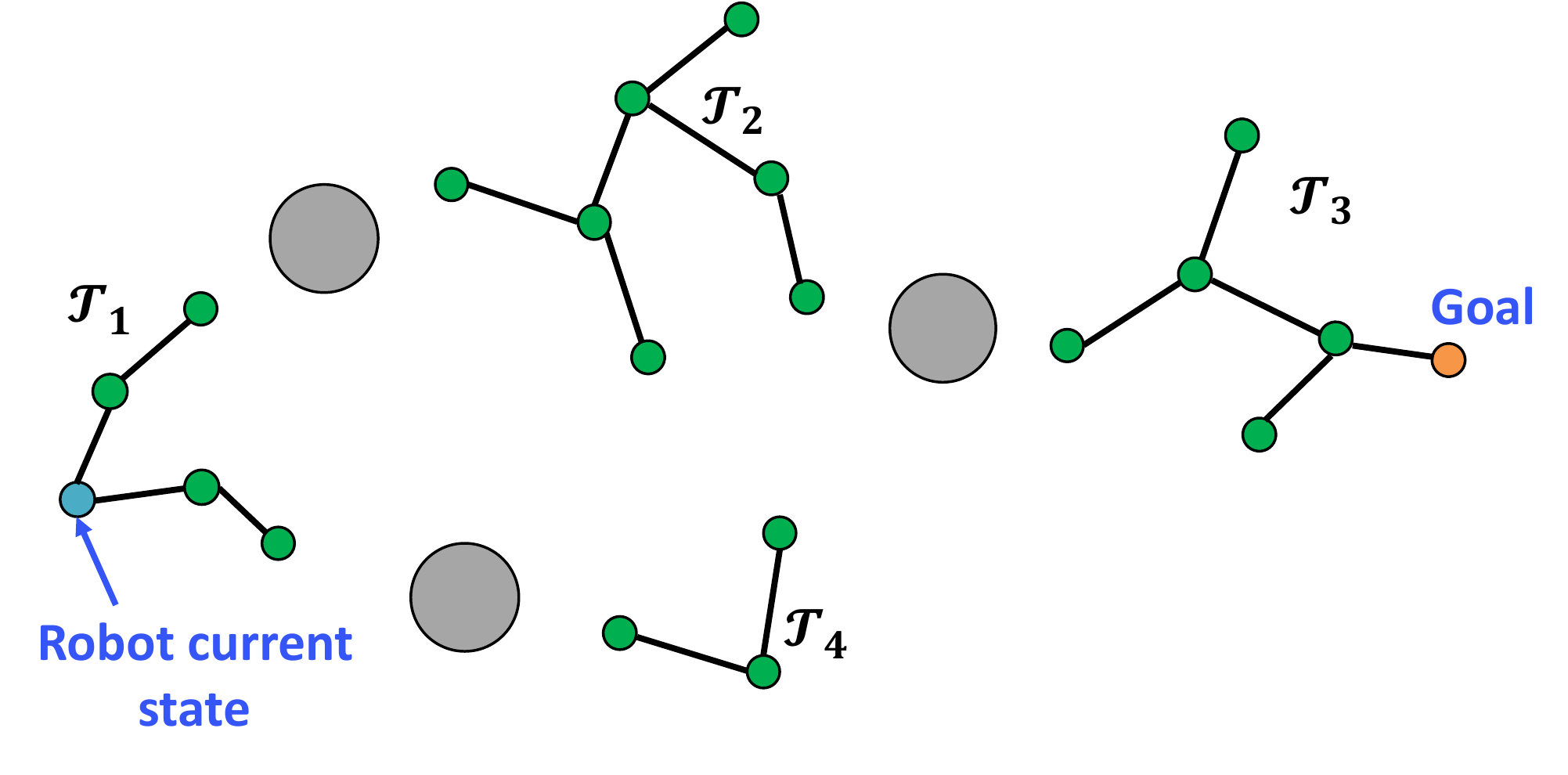}\label{fig:part2}}\\
         \centering
    \subfloat[A random sample is generated; it is joined to all disjoint trees in the local neighborhood by connecting to the nearest node.]{
        \includegraphics[width=0.44\textwidth]{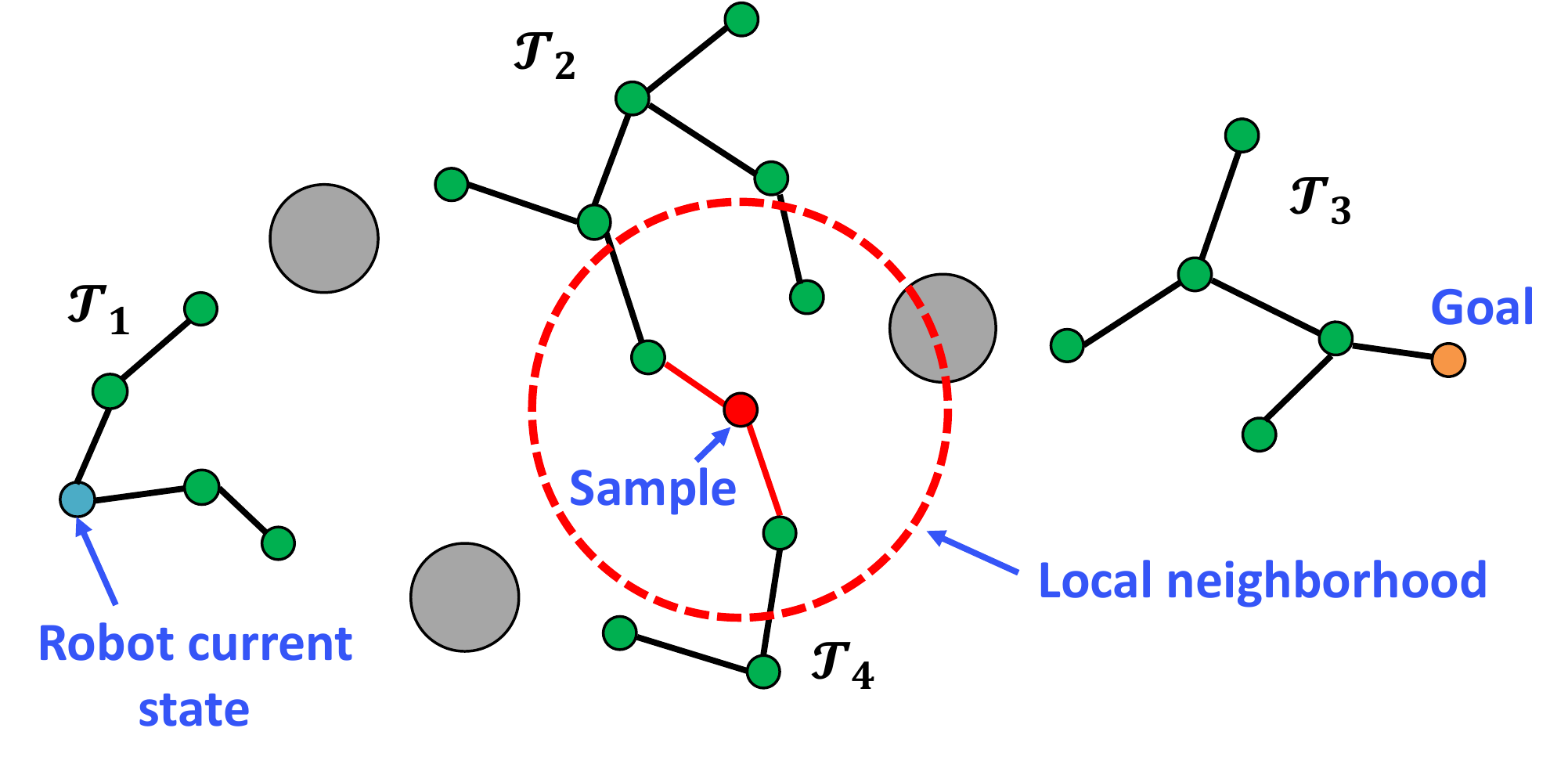}\label{fig:part3}}\quad
         \centering
    \subfloat[A random sample is generated; it is joined to all disjoint trees in the local neighborhood by connecting to the nearest node.]{
        \includegraphics[width=0.44\textwidth]{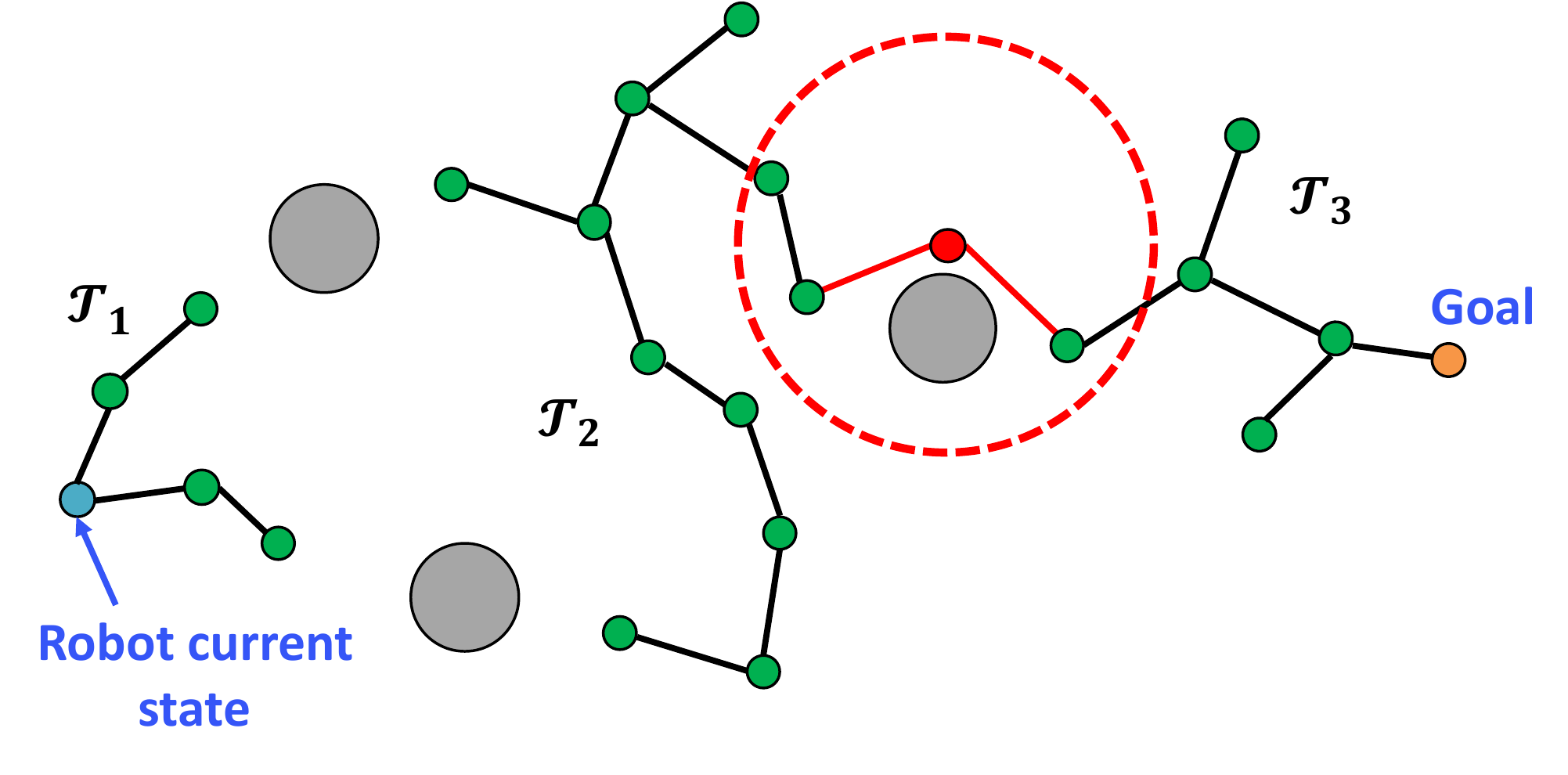}\label{fig:part4}}\\
         \centering
    \subfloat[A random sample is generated; it is joined to all disjoint trees in the local neighborhood by connecting to the nearest node.]{
         \includegraphics[width=0.44\textwidth]{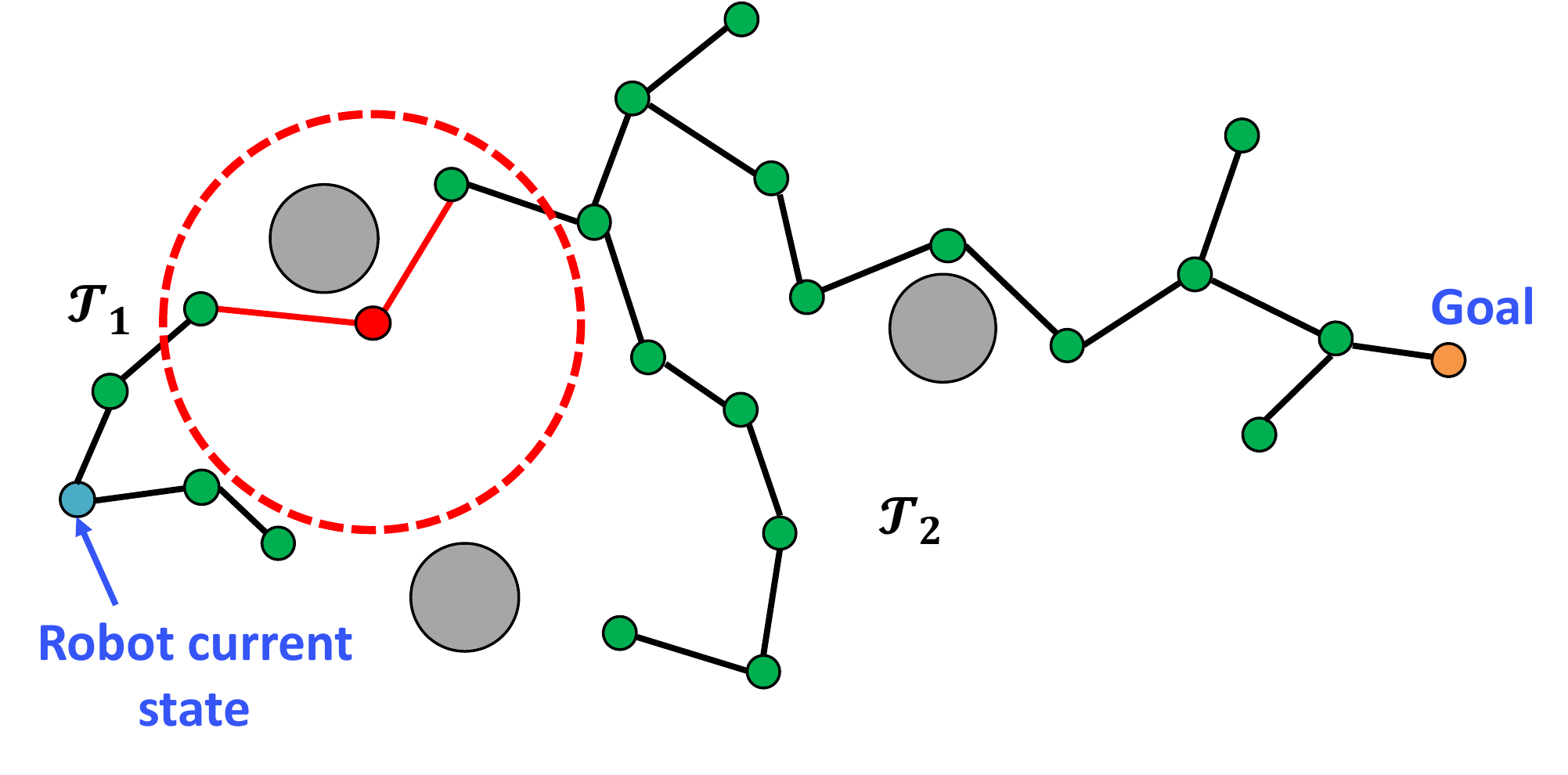}\label{fig:part5}}\quad
         \centering
    \subfloat[Once the full tree is obtained, a new feasible solution is searched on it.]{
        \includegraphics[width=0.44\textwidth]{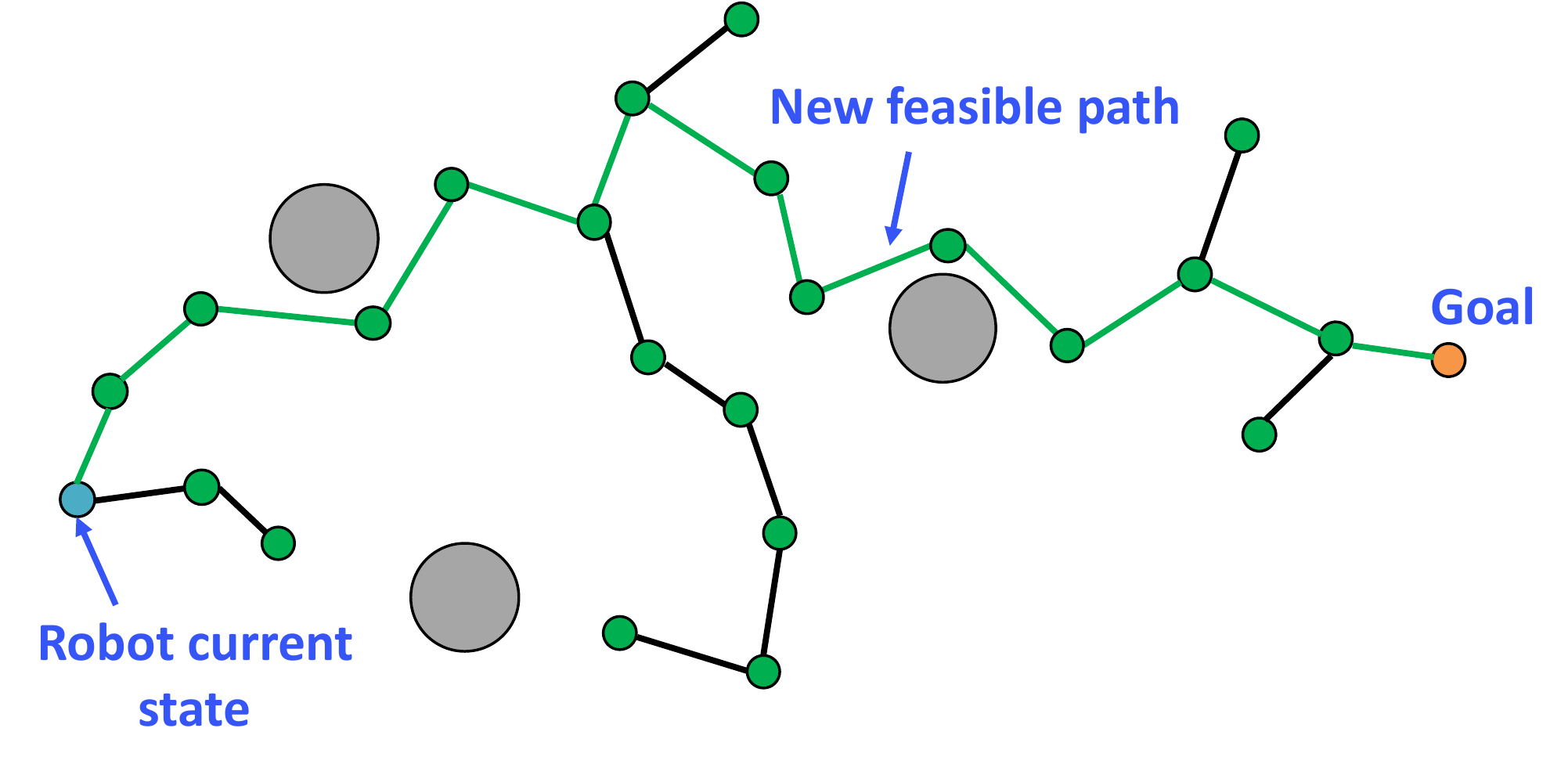}\label{fig:part6}}\\

          \caption{An illustrative example of MRRT algorithm.}\label{fig:example}
          \vspace{-10pt}
\end{figure*}

We present a novel online reactive sampling-based motion planning algorithm for dynamic environments, called Multiple Rapidly Exploring Random Trees (MRRT), to address the limitations of existing approaches. The proposed algorithm is built upon the RRT algorithm and multi-tree structure. At the beginning, RRT algorithm is applied to find the initial solution based on the partial knowledge of the environment. Then, the robot starts to execute this path. At each iteration, the new obstacle configurations are collected by the robot's sensor and used to replan the path. This new information can come from unknown static obstacles (e.g., seafloor layout) as well as moving obstacles. Then, to accommodate the environmental changes, two procedures are adopted: 1) edge pruning, and 2) tree regrowing. Specifically, the edge pruning procedure checks the collision status through the tree and only removes the invalid edges while maintaining the tree structure of already-explored regions. Due to removal of invalid edges, the tree could be broken into multiple disjoint trees. As such, the RRT algorithm is applied to regrow the trees. Specifically, a sample is created randomly and joined to all the disjoint trees in its local neighborhood by connecting to the nearest nodes. Finally, a new solution is found for the robot. Figure~\ref{fig:example} shows an execution example of the online motion planning algorithm in a dynamic environment with moving obstacles.

The advantages of the proposed algorithm are as follows: i) retains the maximal tree structure by only pruning the edges which collide with the obstacles, ii) guarantees probabilistic completeness, and iii) is computational efficient for fast replanning since all disjoint trees are maintained for future connections and expanded simultaneously.

In the full paper, we will present details of the proposed MRRT algorithm and the simulation results of rapid path replanning in an unknown static environment as well as in a dynamic environment containing moving obstacles.


\vspace{-12pt}

\bibliographystyle{IEEEtran}
\bibliography{reference}
\end{document}